\documentclass{article}
\usepackage[square,numbers]{natbib}
\usepackage[utf8]{inputenc} 
\usepackage[T1]{fontenc} 
\usepackage{hyperref} 
\usepackage{url} 
\usepackage{booktabs} 
\usepackage{amsfonts} 
\usepackage{nicefrac} 
\usepackage{microtype} 
\usepackage{lipsum}
\usepackage{fancyhdr} 
\usepackage{graphicx} 
\graphicspath{{media/}} 
\usepackage{xcolor}
\usepackage[para]{threeparttable}

\usepackage{amsmath}
\usepackage{nccmath}
\usepackage{tabularx}
\usepackage{multirow}
\usepackage{amssymb}
\usepackage{subfig}

\usepackage{multicol}
\usepackage{array}
\usepackage{physics}
\usepackage{authblk}

\usepackage{tcolorbox}
\usepackage{longtable}
\graphicspath{ {images/} }
\providecommand{\keywords}[1]
{
  \small	
  \textbf{\textit{Keywords---}} #1
}

\title{ChatGPT: The End of Online Exam Integrity?

}
\author[1]{Teo Susnjak}
\affil[1]{School of Mathematical and Computational Sciences, Massey University, Auckland, New Zealand }
\begin{document}
\maketitle

\begin{abstract} 

    This study evaluated the ability of ChatGPT, a recently developed artificial intelligence (AI) agent, to perform high-level cognitive tasks and produce text that is indistinguishable from human-generated text. This capacity raises concerns about the potential use of ChatGPT as a tool for academic misconduct in online exams. The study found that ChatGPT is capable of exhibiting critical thinking skills and generating highly realistic text with minimal input, making it a potential threat to the integrity of online exams, particularly in tertiary education settings where such exams are becoming more prevalent. Returning to invigilated and oral exams could form part of the solution, while using advanced proctoring techniques and AI-text output detectors may be effective in addressing this issue, they are not likely to be foolproof solutions. Further research is needed to fully understand the implications of large language models like ChatGPT and to devise strategies for combating the risk of cheating using these tools. It is crucial for educators and institutions to be aware of the possibility of ChatGPT being used for cheating and to investigate measures to address it in order to maintain the fairness and validity of online exams for all students.
\end{abstract}



\maketitle

\section{Introduction}\label{sec1}

Higher education has seen a significant shift towards online learning in recent years, and this trend has been accelerated by the COVID-19 pandemic \cite{barber2021}. Many Higher Education Institutions (HEIs) have had to quickly adapt to the challenges posed by the pandemic by transitioning to online classes and exams\cite{butler2020systematic,coghlan2021good,henderson2022factors}. It is unlikely that these trends towards online education will reverse in the near future \cite{dumulescu2021academic} notwithstanding the challenges encountered, since the benefits of remote learning have become appreciated by both HEIs and students alike \cite{barber2021}.

As the sector has increasingly moved online, concerns around academic integrity have also been amplified \cite{barber2021,whisenhunt2022maximizing,garg2022systematic}. The transition to online exams, in particular, has raised concerns about the potential for cheating and other forms of academic misconduct \cite{arnold2016cheating,ahsan2021contract,crook2021internet,noorbehbahani2022systematic,henderson2022factors}. This is due, in part, to the anonymity and lack of direct supervision that are inherent to online exams, as well as the ease with which students may be able to access and share resources during the exam. 

While concerns around academic integrity in online exams have been raised, there is a lack of definitive research with no consolidated literature reviews yet conducted quantifying the extent of dishonest practices in online assessments \cite{garg2022systematic}. 
Indications are that the prevalence is on the rise. In earlier studies, \citet{fask2014online,Corrigan2015,alessio2017examining} detect that significant rates of cheating occurred in online assessments, while in general, \citet{arnold2016cheating} mention that there is a belief among educators that academic misconduct is on the rise and that online assessment is particularly conducive to cheating. More recently, \citet{noorbehbahani2022systematic} reported that cheating in online exams covering more than a decade of research, found that dishonesty in online exams is more prevalent than in traditional face-to-face exams.

To preserve academic integrity in online exams, HEIs have implemented revised recommendations for formulating assessments \cite{whisenhunt2022maximizing}, various technological strategies such as proctored exams \cite{alessio2017examining}, plagiarism detection software, exam security measures, as well as revisions of institutional academic integrity policies and educational campaigns to deter misconduct and honor codes \citet{Corrigan2015}. While these strategies individually or in tandem, may be effective in mitigating the risk of academic misconduct, there is currently insufficient evidence regrading their overall effectiveness in preserving academic integrity in online exams. Meanwhile, ethical concerns surrounding the use of proctoring software on personal computers \cite{alin2022addressing} and their recent challenges on legal grounds \cite{justia2021} have gained momentum.

An additional measure that HEIs have explored is a shift towards using more challenging exam questions \cite{khan2022perspectives} that require greater degrees of critical thinking. \citet{whisenhunt2022maximizing}\footnote{The authors develop a set of recommendations for conducting multiple-choice exams in online environments. Appendix \ref{appendix} demonstrates the ability of ChatGPT to answer such questions.} note that these types of assessments comprising essays and short-answer responses are generally perceived by educators to be more suitable at measuring critical thinking \cite{stanger2012multiple} as well as facilitating deeper learning \cite{watters2007approaches,zheng2008application}. The underlying intention behind them is to move away from multiple-choice and simple information-retrieval questions since these types of questions are regarded as more susceptible to cheating \cite{noorbehbahani2022systematic} when encountering misconduct involving unauthorized web access.

However, a new threat to the academic integrity of online exams, even ones requiring high-order reasoning has emerged. With the recent \footnote{A beta research release of ChatGPT took place on November 30, 2022} public release of ChatGPT by OpenAI\cite{OpenAI}, the world has seen a significant leap in AI capabilities that involve natural language processing and reasoning. This publicly\footnote{The technology was available to the public as of the publication date of this paper} available technology is not only able to engage in sophisticated dialogue and provide information on virtually all topics. It is also able to generate compelling and accurate answers to difficult questions requiring an advanced level of analysis, synthesis, and application of information, as will be demonstrated in this study. It can even devise critical questions itself, the very questions that educators in different disciplines would use for their students' evaluation of competencies. Assuming that high-stakes exams will continue to be perceived as valuable and will continue to be used in education, this development may spell the end of the academic integrity of online examinations. It is therefore imperative that the capabilities of this AI agent be examined. 

\subsection{ChatGPT}
ChatGPT is a large language model. A large language model is a type of AI that uses deep learning (a form of machine learning) to process and generate natural language text. These models are trained on massive amounts of text data, allowing them to learn the nuances and complexities of human language. In the case of ChatGPT, it was trained on a diverse range of text data which included books, articles, and online conversations, to enable it to engage in non-trivial dialogue and provide accurate information on a wide range of topics \footnote{The details of the datasets used to train ChatGPT have not been publicly released.}. The development of ChatGPT represents a significant advancement in the field of natural language processing and AI in general, building upon the initial GPT (Generative Pretrained Transformer\cite{vaswani2017attention}) model and paving the way for further innovations in this area.

One of the key advantages of these large language models is their ability to understand the context of a given prompt and generate appropriate responses. This paper focuses on demonstrating this capability. This is a significant improvement over earlier language models, which were often unable to interpret the meaning and intent behind a given piece of text. Another important aspect is its ability to generate high-quality text that is difficult to distinguish from human writing. 
With its ability to draw out knowledge and answer difficult academic questions, it is inherently capable of answering  examination questions that would otherwise not easily be answered through web searches, and to provide accurate and reliable responses.

\subsection{Aims and Contribution}\label{sec1}

The goal of this article is to provide a first analysis of the capabilities of ChatGPT in demonstrating the complexity of its reasoning and its ability to answer non-trivial university-level questions across multiple disciplines, and its ability to present its responses with clarity and coherence. 

The contribution of this study is its novelty and the urgency of raising an alarm about the unprecedented threat that large language models like ChatGPT pose to academic integrity. The intention is to highlight the fact that current mitigation strategies, recommendations, and processes for preserving academic integrity are likely not yet capable of addressing this danger. 

While this technology is arguably one of the greatest advancements in AI capabilities since the advent of web search engines, and will offer enormous opportunities for productivity advancements, it also brings a fresh challenge to the HEI sector which is already in a state of disruption, but compelled to examine this development with urgency.

\section{Background}\label{sec1}

The literature review examines the most recent investigations into the problem of academic integrity, with a greater focus on the context of online assessments. 

\citet{butler2020systematic} conducted a systematic review that highlighted the transformation of learning and teaching towards more active learning environments, particularly in the context of the COVID-19 pandemic. This has led to the adoption of online examination formats which the authors discuss as being driven by a desire to increase international enrollments, and the 'massification' of higher learning, while the impact of the pandemic has been to accelerate these trends. The authors identify the limitations and challenges of online examinations, including cheating issues, together with access to technology, and the lack of standardized approaches. The study concludes by calling for further research on online examinations and the importance of designing online examinations that are fair, valid, and reliable.

In a comprehensive report, \citet{barber2021} discuss that academic misconduct, including plagiarism and cheating, is a concern for higher education institutions and educators in both in-person and digital assessments. Technology has played a role in helping institutions detect plagiarism, and new developments in technology, such as biometric authentication, authorship analysis, and proctoring software, are being used to identify misconduct. However, the use of proctoring software has also raised concerns about privacy and international students have had issues with the software due to differences in bandwidth. The shift to digital teaching and learning during the COVID-19 pandemic has prompted a review of assessment approaches, including the use of open-book online exams and more authentic, integrated assessments.

\citet{coghlan2021good} report that online exam proctoring technologies, which use AI and machine learning, have gained attention. However, the study notes that these technologies have faced controversy and ethical concerns, including questions about student privacy, potential bias, and the validity and reliability of the software. Some universities have defended their use, while others have retreated from or rejected the use of these technologies. 

A recent systematic literature review by \citet{noorbehbahani2022systematic} on cheating in online exams covering more than a decade of research found that cheating in online exams is indeed a significant concern. The study claims that cheating in online exams is more prevalent than in traditional face-to-face exams. The authors note that a wide range of technologies and tools can be used to facilitate cheating, such as remote desktop and screen sharing, searching for solutions on the internet, and using social networks. Apart from online proctoring, the authors identify a combination of prevention strategies, such as cheat-resistant questions, and detection methods, such as plagiarism detection software and machine learning algorithms, as potentially being effective.

\citet{henderson2022factors} also found that the prevalence of cheating in online exams is a significant issue, while this still remains an issue even in on-campus, paper-based invigilated exams. The authors' findings point to previous research which has shown that cheating persists despite security measures, with conflicting existing evidence about the impact of invigilation and online security on cheating. Their conclusion is that while technology-based security measures can impact student experience and attitudes towards integrity, they do not necessarily reduce cheating.

With respect to proctoring software, \citet{alin2022addressing} also raised the issue of ethical concerns surrounding the use of this technology on personal computers and the interpretation of what constitutes suspicious behavior. The authors stress that even when proctoring systems are permitted for use on online examinations, the exams can still be vulnerable to cheating. The authors posit that there is currently a lack of understanding about how cheating may occur in virtual proctored exams and how to best mitigate it.

Meanwhile, the \citet{khan2022perspectives} also highlighted that proctoring software that required students to keep their cameras on during online examinations was considered stressful and intrusive to privacy by the students, while they also believed cheating would continue irrespective of the measures. The study also suggested using strategies like replacing multiple-choice questions with short-answer questions and employing tighter time limits. While \citet{koh2022shifting} also noted identified one of the key strategies used by educators as teaching transitioned to an online mode was to convert multiple-choice questions to written critical thinking questions, a move which the students reported as leading to their perception that the examinations become harder.

\section{Methodology}\label{sec1}

The methodology for examining the critical and higher-order thinking capabilities of ChatGPT is described here. Three steps were followed listed below and described in more detail in this section.

\begin{enumerate}
    \item Firstly, ChatGPT was asked itself to generate examples of difficult critical thinking questions that involve some scenario, and which target undergraduate students from various disciplines.
    \item Secondly, ChatGPT was then asked to provide an answer to the generated questions. 
    \item Lastly, ChatGPT was asked to critically evaluate the answer given to the question.
\end{enumerate}

\begin{figure}[htb]
\centering\includegraphics[width=0.8\linewidth]{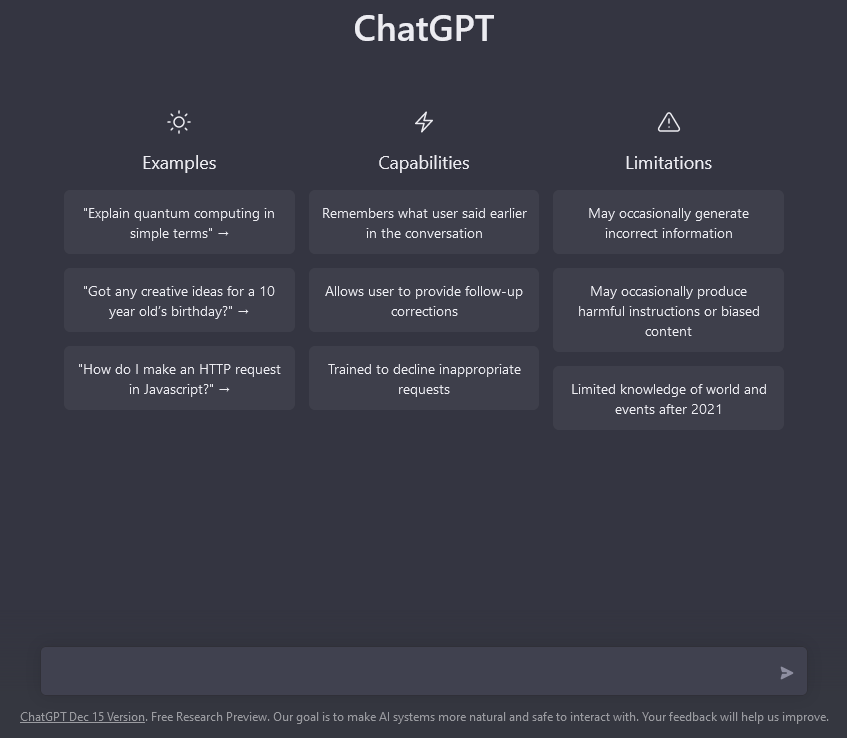}
\caption{\label{fig:chatgpt} The publicly accessible online interface to ChatGPT shows the text input prompt that the bottom.}
\end{figure}

\paragraph{Setup: } An account was created through OpenAI\cite{openaiblog2020} granting access to ChatGPT under the beta research release for experimental usage. Figure \ref{fig:chatgpt} shows the online portal for accessing ChatGPT.

\paragraph{Selected disciplines: } A broad range of discipline areas from the Sciences, Education Studies, Humanities, and Business were selected for demonstrative purposes. Specifically, ChatGPT was prompted to generate subject-specific questions and responses with respect to specific disciplines of Machine Learning, Marketing, History, and Education.

\paragraph{Question generation: } To test the ability of ChatGPT to generate critical thinking questions across multiple disciplines, the model was first provided with a prompt indicating that an example of critical thinking questions for each field of study was sought, which were challenging but appropriate for undergraduate students. This was accomplished by posing the following prompt for each discipline:

\begin{center}
\centering
\begin{tcolorbox}[colback=gray!10,colframe=gray!50,width=10cm]
"Please generate an example of a difficult critical question in the field of Machine Learning targeting undergraduate students which involves an hypothetical scenario."
\end{tcolorbox}    
\end{center}

\paragraph{Answer generation: }  ChatGPT was subsequently prompted to answer the questions it devised for each discipline with specific instructions on the structure and form of the response as follows 

\begin{center}
\centering
\begin{tcolorbox}[colback=gray!10,colframe=gray!50,width=10cm]

"Please answer the following question in several paragraphs using 500 words with examples and supporting arguments."
\end{tcolorbox}    
\end{center}

\paragraph{Critical evaluation of responses: }  To test the ability of ChatGPT to critically evaluate and analyze the responses it has generated, and to provide suggestions for improvement, the model was prompted to consider the original question posed and the answer given with the following additional instruction:

\begin{center}
\centering
\begin{tcolorbox}[colback=gray!10,colframe=gray!50,width=10cm]
"Can you please critically evaluate the following response to the above question, listing its strengths and weaknesses as well as suggestions for improvement."
\end{tcolorbox}    
\end{center}

\subsection{Evaluation of responses}

Universal intellectual standards are standards that a suitable for use in evaluating the quality of reasoning about any problem, issue, or situation \cite{paul1990critical}.
The level of critical thinking expressed ought to consider the purpose, information, and implications of the arguments presented, and these need to be evaluated for quality using universal intellectual standards as outlined by \citet{paul2005state}, with persuasiveness and originality also added. To that end, the following criteria representing dimensions of critical thinking will be applied to the responses to assess the quality of the responses provided by ChatGPT to the prompts outlined above.

    \paragraph{Relevance:} Is the idea being expressed relevant to the topic or question at hand? Does it address the complexities of the issue?

    \paragraph{Clarity:} Is the text easy to understand? Is it well-structured and logically organized? Does it use appropriate language and vocabulary for the intended audience?

    \paragraph{Accuracy:} Is the idea being expressed true or accurate? Can it be verified through evidence or other means?

    \paragraph{Precision:} Is the idea being expressed specific and detailed enough? Is it precise and unambiguous?

    \paragraph{Depth:} Does the idea being expressed go beyond the surface level and consider the underlying complexities and nuances of the issue? Does the text provide a thorough and in-depth analysis of the topic? Does it consider multiple perspectives and present a balanced view?

    \paragraph{Breadth:} Does the idea being expressed consider the full range of relevant perspectives and viewpoints on the issue?

    \paragraph{Logic:} Does the idea being expressed follow logical and consistent reasoning? Are the conclusions supported by the evidence presented?

    \paragraph{Persuasiveness:} Does the text effectively persuade the reader to accept its arguments or conclusions? Is the evidence presented strong and convincing?

    \paragraph{Originality:} Does the text offer new insights or ideas, or does it simply repeat information that is already widely known?

\section{Results}\label{sec1}

The responses by ChatGPT to the prompts outlined in the methodology are shown in four tables, with each one representing a separate discipline, namely Education in Table \ref{educationresponses}, Machine Learning in Table \ref{mlresponses}, History in Table \ref{historyresponses} and Marketing in Table \ref{marketingresponses}.

\begin{table}[p]
\caption{The example of a critical-thinking question generated by ChatGPT and its underlying reasoning for the discipline of Education. The response to the question posed as generated by ChatGPT, together with its own critical evaluation of its answer with suggestions for improvement.}   
\fontsize{7pt}{8pt} 
\selectfont
\begin{tabular} {p{0.3\textwidth} p{0.69\textwidth}@{}}
\hline 
Question with Rationale & Answer
\\
\hline 
"In a hypothetical scenario where a school district is experiencing high rates of student absenteeism, how might implementing a personalized learning program impact attendance rates and student achievement, and what considerations should be taken into account when designing and implementing such a program?" 
\textbf{Rationale:}
\textit{The above question is suitable because it presents an hypothetical scenario that is relevant to the field of education and challenges undergraduate students to think critically about how to address a real-world problem. The question asks students to consider the potential impacts of a personalized learning program on attendance rates and student achievement, as well as the considerations that should be taken into account when designing and implementing such a program. This type of question encourages students to apply their knowledge and understanding of education theories and practices to a hypothetical situation and consider the potential consequences of different approaches. Overall, the question is suitable because it challenges students to think critically and creatively about a relevant issue in the field of education.}
& 
In a hypothetical scenario where a school district is experiencing high rates of student absenteeism, implementing a personalized learning program may have a positive impact on attendance rates and student achievement. Personalized learning programs are designed to tailor instruction to the individual needs and interests of each student, which can increase motivation and engagement in learning. When students feel that their learning is meaningful and relevant to their own lives, they are more likely to attend school regularly and put forth effort in their studies.

There is evidence to support the idea that personalized learning programs can improve attendance rates and student achievement. For example, a study by the U.S. Department of Education found that students in personalized learning programs had higher attendance rates and made greater gains in academic achievement compared to their peers in traditional programs. Similarly, a review of research on personalized learning by the Clayton Christensen Institute found that personalized learning programs had positive impacts on student engagement, motivation, and achievement.

However, there are also considerations that need to be taken into account when designing and implementing a personalized learning program in a school district experiencing high rates of student absenteeism. One consideration is the need for adequate resources and support. Personalized learning programs often require more individualized attention from teachers and may require additional technology and materials. It is important to ensure that the school has the necessary resources and support in place to effectively implement a personalized learning program. Another consideration is the need for proper training and professional development for teachers. Personalized learning approaches may be different from traditional methods, and teachers may need support and guidance to effectively implement these approaches in their classrooms.

Overall, implementing a personalized learning program in a school district experiencing high rates of student absenteeism may have a positive impact on attendance rates and student achievement. However, it is important to carefully consider the resources and support needed to effectively implement such a program, as well as the training and professional development needs of teachers. By taking these factors into account, a school district can increase the chances of success in improving attendance rates and student achievement through personalized learning.
 \\

\hline 

\multicolumn{2}{p{1.02\textwidth} }{\textbf{Critical evaluation of the answer:}

\textbf{Strengths:}     The response provides evidence to support the idea that personalized learning programs can improve attendance rates and student achievement.
    The response acknowledges the need for adequate resources and support as well as proper training and professional development for teachers in order to effectively implement a personalized learning program.

\textbf{Weaknesses:}    The response does not provide specific examples or details about the studies or research mentioned. It would be helpful to have more information about the context and methods of these studies in order to better understand their findings and how they apply to the hypothetical scenario.
    The response does not address any potential challenges or drawbacks of implementing a personalized learning program in a school district experiencing high rates of student absenteeism. It would be helpful to consider potential barriers or difficulties that may arise and how they could be addressed.

\textbf{Suggestions for improvement:}     Provide specific examples and details about the studies or research mentioned, including information about the context and methods used. This would help to provide more context and credibility to the findings.
    Consider potential challenges or drawbacks of implementing a personalized learning program in a school district experiencing high rates of student absenteeism and suggest strategies for addressing them. This would provide a more balanced and nuanced perspective on the potential impacts of personalized learning.
    Consider other factors that may impact attendance rates and student achievement in this hypothetical scenario, such as socio-economic factors, school culture, or community resources. This would provide a more comprehensive analysis of the issue and suggest more holistic approaches to addressing it.
} \\
     \hline
\end{tabular}
\label{educationresponses}
\end{table}

\begin{table}[p]
\caption{The example of a critical-thinking question generated by ChatGPT and its underlying reasoning for the discipline of Machine Learning. The response to the question posed as generated by ChatGPT, together with its own critical evaluation of its answer with suggestions for improvement.}   
\fontsize{7pt}{8pt} 
\selectfont
\begin{tabular} {p{0.3\textwidth} p{0.67\textwidth}@{}}
\hline 
Question with Rationale & Answer
\\
\hline 
"Imagine you are working on a machine learning project to predict the likelihood of a customer churning from a subscription service. Your model is performing well on the training data, but when you test it on unseen data, the accuracy drops significantly. What could be the possible reasons for this discrepancy and how would you troubleshoot the issue?" 
\textbf{Rationale:}
\textit{The above question is suitable because it presents an hypothetical scenario that is commonly encountered in real-world machine learning projects. It challenges the students to think critically about the various factors that can impact the performance of a machine learning model, and to consider how they might go about troubleshooting and addressing any issues. Additionally, the question touches upon the importance of evaluating a model on unseen data, which is a crucial step in the machine learning process. Overall, the question requires the students to apply their knowledge of machine learning principles and techniques to a realistic scenario, which helps to deepen their understanding of the field.}

& 
One possible reason for the discrepancy in performance between the training data and unseen data could be overfitting. Overfitting occurs when a model is too complex and is able to fit the training data extremely well, but is not able to generalize to new data. This can be caused by having too many parameters in the model, or by having a limited amount of training data. In either case, the model becomes too specialized to the training data and is not able to accurately predict outcomes for new data.

To troubleshoot this issue, one approach would be to try simplifying the model by reducing the number of parameters or increasing the amount of training data. This can help the model to more accurately capture the underlying patterns in the data and improve its generalization performance. Another option would be to use techniques such as regularization, which helps to prevent overfitting by introducing a penalty for using too many parameters in the model.

If overfitting is not the cause of the discrepancy, there could be other factors at play. One possibility is that the training data and unseen data come from different distributions. For example, the training data may be collected from a specific time period or geographic region, while the unseen data may come from a different time period or region. In this case, the model may be performing well on the training data because it has learned patterns that are specific to that data, but those patterns may not generalize to the unseen data.

To address this issue, one solution could be to ensure that the training data is representative of the overall distribution of data that the model will be applied to. This can be achieved through sampling techniques such as stratified sampling, which ensures that the training data contains a balanced representation of different subgroups within the overall data. Another option would be to use domain adaptation techniques, which help the model to learn to adjust to changes in the data distribution.

Overall, there are many potential reasons for a discrepancy in performance between training data and unseen data, and the appropriate approach to troubleshooting will depend on the specific circumstances of the project. By carefully evaluating the model and the data, and applying appropriate techniques to address any issues, it is possible to improve the generalization performance of the model and increase its accuracy on new data.
 \\

\hline 

\multicolumn{2}{p{1.0\textwidth} }{\textbf{Critical evaluation of the answer:}

\textbf{Strengths:}
    The response correctly identifies overfitting as a possible cause of the discrepancy in performance between the training data and unseen data.
    The response provides several suggestions for troubleshooting the issue, including simplifying the model, using regularization techniques, and ensuring that the training data is representative of the overall data distribution.

\textbf{Weaknesses:}
    The response does not mention other possible causes of the discrepancy beyond overfitting and data distribution, such as bias in the training data or a lack of relevant features in the model.
    The response does not provide specific examples of techniques that can be used to simplify the model or ensure that the training data is representative of the overall data distribution.

\textbf{Suggestions for improvement:}
    Mention other potential causes of the discrepancy, such as bias in the training data or a lack of relevant features in the model.
    Provide specific examples of techniques that can be used to simplify the model or ensure that the training data is representative of the overall data distribution, such as feature selection or sampling techniques.
    Explain the benefits and drawbacks of each technique suggested, and provide guidance on how to decide which technique is most appropriate for a given situation.
} \\
     \hline
\end{tabular}
\label{mlresponses}
\end{table}

\begin{table}[p]
\caption{The example of a critical-thinking question generated by ChatGPT and its underlying reasoning for the discipline of History. The response to the question posed as generated by ChatGPT, together with its own critical evaluation of its answer with suggestions for improvement.}   
\fontsize{7pt}{8pt} 
\selectfont
\begin{tabular} {p{0.3\textwidth} p{0.67\textwidth}@{}}
\hline 
Question with Rationale & Answer
\\
\hline 

"Imagine that a new discovery has been made that completely changes our understanding of a major historical event. How would this discovery impact our understanding of the past, and what steps would need to be taken to incorporate it into the historical narrative?" 

\textbf{Rationale:}

\textit{This question requires a deep understanding of the role of historical narratives in shaping our understanding of the past, as well as the ability to think critically about the potential impacts and implications of a major historical discovery. It also requires an awareness of the ways in which different perspectives and interpretations can impact our understanding of history, and the ability to propose potential solutions to these challenges in the context of a hypothetical scenario.}

& A new discovery that completely changes our understanding of a major historical event would have a profound impact on our understanding of the past. This discovery would likely challenge long-held beliefs and assumptions about the event in question, and would require historians to reevaluate and revise their interpretations of the past.

One example of this type of discovery might be the uncovering of new evidence about a key event in world history, such as the assassination of President John F. Kennedy. For decades, the assassination has been viewed as the work of a lone gunman, but imagine if new evidence were discovered that indicated a broader conspiracy was at play. This new information would fundamentally change our understanding of the event and its place in history.

To incorporate this discovery into the historical narrative, a number of steps would need to be taken. First, historians would need to carefully review and evaluate the new evidence to determine its authenticity and relevance. This process might involve conducting additional research and analysis, as well as seeking the input of other experts in the field. If the evidence is found to be credible, it would then be necessary to incorporate it into the existing historical narrative.

This process would involve revising existing interpretations of the event and developing new theories to explain the new information. This might involve reexamining other historical events and sources in light of the new discovery, as well as considering how the discovery changes our understanding of the broader historical context. In addition, it would be important to communicate the new discovery to the public and other stakeholders, such as educators and policymakers, to ensure that the revised historical narrative is widely understood and accepted.

Overall, a new discovery that completely changes our understanding of a major historical event would have significant implications for our understanding of the past. By carefully evaluating and incorporating this new information into the historical narrative, we can ensure that our understanding of the past is accurate and complete. \\

\hline 

\multicolumn{2}{p{1.0\textwidth} }{\textbf{Critical evaluation of the answer:}

The response to the question is a well-written and detailed discussion of the potential impact of a new discovery that completely changes our understanding of a major historical event. The response provides a clear explanation of how such a discovery would challenge existing beliefs and interpretations of the past, and offers a thoughtful analysis of the steps that would need to be taken to incorporate the discovery into the historical narrative.

One \textbf{strength} of the response is its use of a specific example to illustrate the potential impact of a new discovery. By discussing the hypothetical discovery of new evidence about the assassination of President John F. Kennedy, the response provides a concrete illustration of how a new discovery might challenge existing historical narratives and require historians to revise their interpretations of the past.

Another strength of the response is its thoroughness. The response covers all of the key steps that would need to be taken to incorporate a new discovery into the historical narrative, from carefully evaluating the evidence to revising existing interpretations and communicating the new information to the broader public.

One potential \textbf{weakness} of the response is that it does not discuss the potential challenges and difficulties that might arise in the process of incorporating a new discovery into the historical narrative. For example, the response does not mention the potential for resistance or skepticism from other historians or members of the public who may be attached to the existing historical narrative. Additionally, the response does not address the potential impact of the new discovery on broader issues, such as the teaching of history or public policy.

\textbf{To improve the response}, one suggestion would be to provide more detail and analysis on the potential challenges and difficulties that might arise in the process of incorporating a new discovery into the historical narrative. This could include discussing the potential for resistance or skepticism from other historians or members of the public, as well as the potential impact of the new discovery on broader issues such as the teaching of history or public policy. Additionally, the response could be strengthened by providing more concrete examples of how a new discovery might challenge existing historical narratives and require historians to revise their interpretations of the past.
} \\

     \hline
\end{tabular}
 
\label{historyresponses}
\end{table}

\begin{table}[p]
\caption{The example of a critical-thinking question generated by ChatGPT and its underlying reasoning for the discipline of Marketing. The response to the question posed as generated by ChatGPT, together with its own critical evaluation of its answer with suggestions for improvement.}   
\fontsize{7pt}{8pt} 
\selectfont
\begin{tabular} {p{0.3\textwidth} p{0.7\textwidth}@{}}
\hline 
Question with Rationale & Answer
\\
\hline 
"Imagine you are a marketing manager at a company that sells organic skincare products. A competing company has just released a new line of synthetic skincare products that are significantly cheaper and more widely available in stores. Your target audience is primarily health-conscious millennial women. How would you develop a marketing strategy to differentiate your organic products and convince consumers to choose your brand over the cheaper synthetic options?" 

\textbf{Rationale:}

\textit{This is a suitable question because it presents an actual scenario that marketing managers may face in the real world and requires the students to think critically about how to approach the problem. It also requires the students to consider the target audience and their motivations, as well as the strengths and weaknesses of both the organic and synthetic products. The question also allows for multiple potential solutions, encouraging the students to consider a variety of strategies and consider the potential outcomes of each option. Overall, this question challenges the students to apply their knowledge and skills to a realistic and complex problem in the field of marketing.}
& 
As a marketing manager at a company that sells organic skincare products, facing competition from a cheaper and more widely available synthetic option can be a daunting challenge. However, with a well-planned and targeted marketing strategy, it is possible to differentiate our organic products and convince consumers to choose our brand over the synthetic options.

One potential approach could be to focus on highlighting the benefits of using organic products. Many consumers, particularly millennial women, are increasingly concerned about the impact of their purchasing decisions on the environment and their own health. By emphasizing the sustainability and natural ingredients of our products, we can appeal to this group's values and differentiate ourselves from the synthetic competition. For example, we could use social media campaigns and influencer partnerships to showcase the benefits of using organic products and the negative consequences of synthetic ingredients on the environment and personal health.

In addition, we can leverage our unique selling points (USPs) to further differentiate ourselves from the competition. For example, if our products are made with rare or hard-to-find organic ingredients, or if we offer a wider range of products than the synthetic brand, we can emphasize these points in our marketing efforts. By focusing on what makes our brand stand out, we can convince consumers that our products are worth the extra cost.

Another strategy could be to target a niche market within the larger group of health-conscious millennial women. For example, we could focus on pregnant women or women with sensitive skin, who may be particularly concerned about the ingredients in their skincare products. By tailoring our marketing efforts to a specific subset of the target audience, we can effectively communicate the benefits of our products to a more targeted group of consumers.

Finally, it may also be beneficial to address consumer concerns about the cost of our products head-on. One way to do this could be through the use of loyalty programs or special offers, such as discounts for first-time buyers or bundle deals. By offering value to our customers, we can help offset the perceived higher cost of our products and make them more appealing to price-sensitive consumers.

Overall, there are many potential approaches to differentiating our organic skincare products and convincing consumers to choose our brand over the cheaper synthetic options. By highlighting the benefits of using organic products, leveraging our USPs, targeting a niche market, and addressing cost concerns, we can effectively communicate the value of our products and convince consumers to make the switch.
 \\

\hline 

\multicolumn{2}{p{1.03\textwidth} }{\textbf{Critical evaluation of the answer:}

The above response to the question presents several potential strategies for differentiating organic skincare products and convincing consumers to choose them over cheaper synthetic options. One \textbf{strength} of the response is that it considers the motivations and values of the target audience, specifically health-conscious millennial women, and suggests targeting marketing efforts towards those values. It also highlights the importance of leveraging unique selling points and addressing cost concerns as ways to differentiate the organic products.

However, there are a \textbf{few weaknesses} in the response that could be addressed for improvement. One weakness is that the response does not consider the potential challenges or drawbacks of implementing the suggested strategies. For example, targeting a niche market may be more effective in the short term, but it may also limit the brand's long-term growth potential. Additionally, offering discounts or loyalty programs may help attract price-sensitive consumers, but it could also erode the perceived value of the products.

Another weakness is that the response does not consider the potential competition from the synthetic brand. For example, the synthetic brand may also target health-conscious consumers or offer loyalty programs of their own. By considering the competitive landscape, the marketing manager can better anticipate the actions of the competition and adjust their strategy accordingly.

Overall, the response presents several potential strategies for differentiating organic skincare products and convincing consumers to choose them over cheaper synthetic options. However, \textbf{to improve the response}, it would be helpful to consider the potential challenges and drawbacks of each strategy and the competitive landscape.
} \\
     \hline
\end{tabular}
\label{marketingresponses}
\end{table}

\subsection{Response Analysis}

There are several key factors to consider when conducting a higher order critical analysis of the regenerated responses.

    \paragraph{Clarity:} Across all responses to the prompts, ChatGPT has demonstrated strong clarity. The language used in the responses is straightforward to understand and follows the structure and conventions of what one would expect from natural language responses. The responses are well-organized and coherent there is an intentional flow of ideas in longer texts. Clarity is also expressed in the rationale provided for the questions and well as in the critical evaluations. The vocabulary, especially the technical language when necessary, and grammar can be regarded as appropriate for the intended audience.

    \paragraph{Accuracy:} In order to fully consider the accuracy of the responses to the questions, evaluations of subject experts from each of the four disciplines would need to be sought. The author can attest to the accuracy of the response and the question and the subsequent critique of the responses with respect to Machine Learning, where the concept of overfitting is well described and examples of techniques that can be used to address it are accurately provided.
    
    It is beyond the scope of this study to draw in subject experts from Marketing, Education (specializing in the U.S. context), and History to assess these responses for accuracy. Returning to the Machine Learning question, the posed question is identical to a question that the author has used in Data Science courses; however, with a different scenario. The justification for generating this question is also correct. The critique of the generated response is astute, and if integrated into the actual exam response, would carry full marks which actual students rarely achieve in the experience of the author.

    \paragraph{Precision:} The responses to the questions are specific as well as detailed. In the context of the Machine Learning response, specific examples of techniques that can be used to troubleshoot the issue were provided. The responses also clearly distinguish between different potential causes of the discrepancy. In the context of Education, the e U.S. Department of Education was drawn into the response as well as the Clayton Christensen Institute. For History, the example of the assassination of President John F. Kennedy was discussed, while the Marketing responses identified specific target groups. The precision in responses is also demonstrated across all critical evaluations where specific and detailed points are provided. 

    \paragraph{Relevance:} In general, for each set of requests for each discipline, the responses provided to the requests to generate an initial exam question, followed by an answer and subsequently a critical evaluation of the answer, are all demonstrably relevant to the prompts. All responses were on-topic and relevant to both the subject matter concerning each discipline and to the intent of the requests, which required the generation of a difficult question involving a hypothetical scenario, followed by an actual answer, and then a critical analysis of the answer.

    \paragraph{Depth:} ChatGPT has demonstrated a noteworthy level of depth in terms of the complexity of the questions formulated, as well as the rationales offered to support them. The answers were constrained to 500 words, and within that constraint, well-thought-out strategies and examples were provided for all disciplines. Meanwhile, the critiques of the responses were substantial, with strengths and weaknesses, together with suggested improvements offered as requested, demonstrating the AI agent's thorough and an in-depth analysis understanding of each of the topics.

    \paragraph{Breadth:} Again, given the constraints within which the responses needed to be given, the provided answers provided explanations of two scenarios in each case. While the suggestions for improvements in each case offered further examples of the agent's capacity for breadth. 
    
    \paragraph{Logic:} All responses follow logical and consistent reasoning, providing specific examples and explanations. The logical coherence is particularly emphasized in rationales given to justify the generation of each question, as well as in the answers to the same questions where different scenarios were explained and well-organized. 

    \paragraph{Persuasiveness:} While subject experts may offer additional points of critique to the answers offered by the AI agent, a remarkable feature of the responses is the confidence with which they are expressed. Arguments and evidence are presented in a clear and logical manner and efforts to address potential counterarguments or objections are mentioned. Ultimately, the quality of the persuasiveness of the responses will vary depending on the perspective of the reader, it is self-evident that they are, rightly or wrongly, expressed without reservations.

    \paragraph{Originality:} The AI agent has been trained on a large dataset of human-generated text that contains a vast amount of accumulated knowledge, and by design, generates responses based on the patterns and conventions found in that dataset. It is therefore hard to conceive that the responses can be original if they are correct. However, since accuracy is a known weak point of large language models, it is conceivable that some connections between topics and events from the dataset corpus made during the training of the models are incorrect (referred to as "hallucinations"), in which case, some responses could be original if they are incorrect. In this scenario, it is probable that the responses provided present information and insights that are based on established knowledge and practices in the field of language models and their applications, and while the responses may not necessarily be original or novel in the sense of presenting completely new ideas, they do seemingly useful information and insights that are sufficient for answering undergraduate examinations.

\section{Discussion}

It is clear from the experimental evidence conducted in this paper that AI technologies have reached exceptional levels and are now capable of critical thinking rather than just information retrieval. The generated responses can be assessed as being clear in exposition, precise with respect to examples used, relevant to the requests while being sufficiently deep and broad considering the constraints imposed while being logically coherent in longer texts. These advances in AI technology are not just incremental improvements, but can arguably be regarded as the most significant technological leap in the last 25 years.  

One of the most impressive capabilities of ChatGPT is its ability to reason critically, as well as express thoughts and ideas in flawless prose. This technology has demonstrated exceptional competency in these areas, seemingly matching the capabilities of humans. \citet{paul2005state} states that a person is a critical thinker to the extent that they regularly learn and critique their thinking in order to improve it. The AI agent's ability to critique its own responses, with clear discussions of its merits and weaknesses, and suggested improvements demonstrates that the agent possesses this capability. The AI agent has exceeded mere memorization and information retrieval capabilities afforded by current web search engines, and is instead now able to conceptualize and demonstrated higher order thinking. 

The full impact of these advances on society will become apparent over time, but the consequences for the education sector are profound and particularly relevant and immediate to online examination integrity.  There is a risk that students could potentially use ChatGPT to cheat on exams, as the technology is able to generate responses that are indistinguishable from those of a human. Experimental findings in this study support this concern. This raises serious questions about the reliability and validity of online exams going forward and the potential for cheating to occur.

Overall, ChatGPT's competency in thinking and reasoning critically, as well as expressing thoughts and ideas in quality prose, presents both opportunities and challenges. While it has the potential to facilitate the exchange of ideas, increase productivity and facilitate learning, it also presents challenges for the education sector. This technology will require careful consideration and oversight, as well as consensus on what constitutes ethical and responsible use of this technology.

\subsection{Recommendations for mitigating strategies}

Finally, when ChatGPT was posed the question: "How would you suggest that students are prevented from using ChatGPT from cheating in online university exams?", the following response was given:

\begin{center}
\centering
\begin{tcolorbox}[colback=gray!10,colframe=gray!50,width=12cm]

"There are a few steps that universities could take in order to prevent students from using ChatGPT or any other AI tool to cheat on online exams.

Firstly, universities could implement measures to prevent students from using ChatGPT or any similar tool during the exam. For example, they could use proctoring software to monitor students' screens and ensure that they are not using any unauthorized tools or resources during the exam. They could also use secure browsers or other technologies to prevent students from accessing external websites or applications during the exam.

Secondly, universities could also implement measures to detect and deter students who attempt to use ChatGPT or any other AI tool to cheat on the exam. For example, they could use plagiarism detection software to identify and flag any exam answers that are suspiciously similar to answers generated by ChatGPT or any other AI tool. They could also use machine learning algorithms to analyze the exam answers and identify any patterns or anomalies that may indicate the use of unauthorized tools or resources.

Thirdly, universities could also educate students about the ethical implications of using ChatGPT or any other AI tool to cheat on exams. They could provide students with information about the university's academic integrity policies, and could remind them of the consequences of using unauthorized tools or resources during the exam. This could help to prevent students from attempting to use ChatGPT or any similar tool to cheat on the exam, and could encourage them to approach the exam honestly and ethically.

In addition to the steps mentioned above, universities could also consider using AI to help prevent students from using ChatGPT or any other AI tool to cheat on online exams. For example, they could use machine learning algorithms to analyze the exam answers and identify any patterns or anomalies that may indicate the use of unauthorized tools or resources. They could also use natural language processing (NLP) algorithms to detect any answers that are suspiciously similar to answers generated by ChatGPT or any other AI tool.

By using AI in this way, universities could improve their ability to detect and deter students who attempt to use ChatGPT or any other AI tool to cheat on the exam. This could help to ensure the fairness, validity, and integrity of the examination process, and could prevent students from gaining an unfair advantage over their peers.

Additionally, universities could also consider using AI to help develop more secure and robust online exam platforms. For example, they could use AI to improve the security of the exam platform, and to prevent students from accessing unauthorized tools or resources during the exam. They could also use AI to improve the user experience of the exam platform, and to make it easier for students to take the exam from any location.

Overall, there are several ways in which universities could use AI to help prevent students from using ChatGPT or any other AI tool to cheat on online exams. By using AI to improve the security and reliability of the exam platform, and to detect and deter cheating, universities can help to ensure the fairness, validity, and integrity of their examination processes."
\end{tcolorbox}    
\end{center}

Perhaps some small signs of weakness of ChatGPT's responses are found in the above, and in fairness, possibly more due to the difficulty of devising a solution to the stated problem than to its capabilities. Firstly, the suggestion of using plagiarism detection software is unlikely to be effective, as ChatGPT is able to generate unique answers given the same prompts - this is particularly the case with small perturbations to the inputs. This means that the answers generated by ChatGPT may not be similar to any other answers, making it difficult for plagiarism software to identify them as suspicious.

Secondly, the suggestion of using machine learning algorithms and natural language processing (NLP) algorithms to detect suspicious answers is also in line with what is currently being used by some HEIs \cite{garg2022systematic}, but may not be realistic at this point if these solutions need to be developed. This solution is likely too expensive and unaffordable for many universities. These technologies require significant resources and expertise to implement and maintain, and may not be feasible for most institutions. There are some indications that GPT-text output detectors already in existence\footnote{ https://huggingface.co/openai-detector/ } have some potential to identify AI-generated text due to an underlying signature in all the text. However, these tools need further research.

With respect to the suggestion of using AI to improve the security and reliability of the online exam platform and to detect and deter cheating may be ineffective. While AI can be a useful tool in these areas, it is not a panacea and may not be able to fully address the issue of cheating using ChatGPT or any other AI tool. Unfortunately, it does not even appear to be effective to ask ChatGPT if it has generated specific pieces of text in order to catch cheating, as preliminary attempts using this strategy have shown that it does not retain records of the text it has generated in previous sessions.

Thirdly, the suggestion of educating students about the ethical implications of using ChatGPT or any other AI tool to cheat on exams is unlikely to be effective in preventing cheating. While education is an important part of promoting academic integrity, it may not be sufficient on its own to deter students who are determined to cheat, and such initiatives have already been shown to only be marginally effective \cite{Corrigan2015}.

One well-known limitation of ChatGPT is its uni-modal input capabilities - meaning that ChatGPT can only accept human text as input. Therefore, in online examinations without effective proctoring software is to be conducted, examinations would need to incorporate more than just text for posing questions. Therefore, the following strategies can be considered:

\begin{itemize}
    \item 
    Use multi-modal channels for exam questions: Embedding images to exam questions can make it more difficult for students to cheat and for ChatGPT to generate accurate responses, as the technology relies on text input only.
    
    \item 
    Experiment with pre-recorded video recorded questions that combine verbal questions with images: This can add an additional layer of difficulty for students attempting to cheat and make it more challenging for ChatGPT to generate accurate responses.

    \item GPT output detection: Check responses against GPT language detector models online at various portals\footnote{https://huggingface.co/openai-detector/}

    \item 
    Return to oral exams: Requiring students to demonstrate their knowledge verbally in real-time online or on-campus premises.

\end{itemize}

It is only a matter of time before large language models evolve into more general AI agents with the ability to incorporate multiple channels\footnote{GPT-4 is under development and multi-modal input capability is reportedly being researched.}, including images, videos, and audio inputs. For now, exploiting the limitations of the technology is the only way to stay ahead, while carefully evaluating the effectiveness of different strategies and continually adapting and refining them as needed.

\subsection{Limitations}

This is a preliminary investigation into the capabilities of ChatGPT to be used for answering critical thinking questions in online settings. As such, further improvements could be made, such as using independent subject experts from Education, History, and Marketing to evaluate the responses, indeed, subject experts and previous exam questions from various courses could be used instead in future studies. However, this study has also demonstrated that ChatGPT is capable of generating effective questions.

\section{Conclusion}

This study has investigated the capability of a recently released AI agent, ChatGPT, to perform high-order thinking tasks and to generate text that is indistinguishable from that of humans, and which could be used as a tool for academic dishonesty in online examinations. The AI agent was prompted to generate questions, to provide and rationale, followed by an answer as well as a critique. 

The study has found that the emergence of technologies like ChatGPT presents a significant threat to the integrity of online exams, particularly in the context of tertiary education where online exams are becoming increasingly common. These models demonstrate a high degree of critical thinking and are able to generate highly realistic text with little input, making it possible for students to cheat on exams.

A return to invigilated and oral exams and the use of advanced proctoring tools may be effective in combating this threat, though they are not a perfect solution. New AI and  machine learning tools capable of detecting text outputs from ChatGPT-like models need to be researched. While further research is needed to fully understand the implications of these large language models and to develop strategies for addressing the potential for cheating using these tools. It is important for educators and institutions to be aware of the potential of this tool to facilitate cheating and to take steps to combat it, in order to maintain the integrity of online exams and ensure fair and valid assessments for all students.


\appendix

\section{Examples of Multiple-choice questions and answers generated by ChatGPT}\label{appendix}

The following are responses to the requests to generate examples of multiple-choice questions across four disciplines considered in this study, and to provide answers with associated explanations.

\subsection{Machine Learning}

\paragraph{Question:} Which of the following is NOT a disadvantage of using a support vector machine (SVM) model for classification tasks?

\begin{itemize}
\item A) SVMs can be sensitive to the choice of kernel and hyperparameters, which can be difficult to select.

\item B) SVMs can have poor generalization performance when the data is not linearly separable.

\item C) SVMs are not sensitive to the scaling of the input features.

\item D) SVMs can be slow to train on large datasets.

\item E) SVMs can be prone to overfitting when the number of features is much greater than the number of training examples.
\end{itemize}

Answer: C) SVMs are not sensitive to the scaling of the input features.

\paragraph{Explanation:} SVMs are a type of linear classifier that seeks to find the hyperplane in feature space that maximally separates the classes. One disadvantage of SVMs is that they can be sensitive to the choice of kernel and hyperparameters, which can be difficult to select and can impact the model's performance. SVMs can also have poor generalization performance when the data is not linearly separable, and they can be slow to train on large datasets. In addition, SVMs can be prone to overfitting when the number of features is much greater than the number of training examples. However, SVMs are not sensitive to the scaling of the input features, meaning that they can perform well even if the features are on different scales.


\subsection{Education}

\paragraph{Question:}
Which of the following theories best explains the process of learning through observation and imitation?

\begin{itemize}
\item A) Piaget's cognitive development theory
\item B) Vygotsky's sociocultural theory
\item C) Bandura's social learning theory
\item D) Bloom's taxonomy of educational objectives
\end{itemize}

Answer: C) Bandura's social learning theory

\paragraph{Explanation:} Bandura's social learning theory explains the process of learning through observation and imitation by stating that individuals learn by watching and mimicking the actions of others. This is known as vicarious learning or modeling. According to the theory, individuals do not only learn new behaviors through direct reinforcement, but also through observing and imitating the actions of others. For example, a child may learn to tie their shoes by watching their parent demonstrate the action and then practicing it themselves. Piaget's cognitive development theory focuses on the child's mental development and the stages they go through in order to understand the world around them. Vygotsky's sociocultural theory emphasizes the role of cultural and social factors in learning and development. Bloom's taxonomy of educational objectives is a classification system that outlines the different levels of cognitive thinking and learning objectives. None of these theories specifically address the process of learning through observation and imitation.


\subsection{Marketing}

\paragraph{Question:}    In which of the following marketing strategies is the company attempting to persuade consumers to purchase their product through emotional appeals and aspirational messaging?

\begin{itemize}
    \item A) Price skimming
\item B) Psychological pricing
\item C) Market segmentation
\item D) Lifestyle branding
\item E) Product positioning
\end{itemize}

Answer: D) Lifestyle branding

\paragraph{Explanation:} Lifestyle branding is a marketing strategy that focuses on creating an emotional connection with consumers by presenting the company's products as a way to enhance the consumer's desired lifestyle or image. This strategy often involves aspirational messaging and appeals to the consumer's emotions in order to persuade them to purchase the product. The other options do not involve emotional appeals or aspirational messaging, so they are not the correct answer.


\subsection{History}

\paragraph{Question:} Which of the following best explains the significance of the Indian Mutiny of 1857 to the British Empire?

\begin{itemize}
    \item A) It marked the beginning of the end of British colonial rule in India.
\item B) It led to the establishment of the British Raj, a period of direct British rule in India.
\item C) It sparked a global debate about the morality of imperialism.
\item D) It resulted in the abolition of the East India Company and the transfer of power to the British Crown.
\item E) It contributed to the development of the British industrial revolution.

\end{itemize}

Answer: D) It resulted in the abolition of the East India Company and the transfer of power to the British Crown.

\paragraph{Explanation:} The Indian Mutiny of 1857, also known as the Sepoy Mutiny, was a significant event in the history of British India. It was a widespread rebellion against the British East India Company, which was the main governing body in India at the time. The mutiny began as a protest against the use of animal fat in the greasing of rifle cartridges, which was offensive to Hindus and Muslims, but it quickly escalated into a broader uprising against British rule. The mutiny was eventually suppressed, but it had far-reaching consequences for the British Empire. One of the most significant outcomes of the mutiny was the abolition of the East India Company and the transfer of power to the British Crown. This marked a major shift in the governance of India, as the British government took direct control of the country and established the British Raj, a period of direct British rule that lasted until 1947.

\end{document}